\documentclass[letterpaper]{article} % DO NOT CHANGE THIS
\usepackage{aaai2026}
\usepackage{times}  % DO NOT CHANGE THIS
\usepackage{helvet}  % DO NOT CHANGE THIS
\usepackage{courier}  % DO NOT CHANGE THIS
\usepackage[hyphens]{url}  % DO NOT CHANGE THIS
\usepackage{graphicx} % DO NOT CHANGE THIS
\usepackage{natbib}  % DO NOT CHANGE THIS AND DO NOT ADD ANY OPTIONS TO IT
\usepackage{caption} % DO NOT CHANGE THIS AND DO NOT ADD ANY OPTIONS TO IT
\usepackage{algorithm}
\usepackage{algorithmic}

\usepackage{newfloat}
\usepackage{listings}
\DeclareCaptionStyle{ruled}{labelfont=normalfont,labelsep=colon,strut=off} % DO NOT CHANGE THIS
\floatstyle{ruled}
\newfloat{listing}{tb}{lst}{}
\floatname{listing}{Listing}
\usepackage{times}
\usepackage{helvet}
\usepackage{courier}
\usepackage{xcolor}

\usepackage{microtype}
\usepackage{tcolorbox}
\tcbuselibrary{breakable}
\usepackage{multirow}
\usepackage{booktabs}
\usepackage[table]{xcolor}
\usepackage{amsmath,amssymb}
\renewcommand{\algorithmiccomment}[1]{\hfill// #1}

\title{Incorporating Self-Rewriting into Large Language Model Reasoning Reinforcement}

\author{
    Jiashu Yao\textsuperscript{\rm 1}\thanks{The research and development work was carried out during Jiashu Yao’s internship at ByteDance China.},
    Heyan Huang\textsuperscript{\rm 1},
    Shuang Zeng\textsuperscript{\rm 3},
    Chuwei Luo\textsuperscript{\rm 3},
    Wangjie You\textsuperscript{\rm 3},
    Jie Tang\textsuperscript{\rm 3},
    Qingsong Liu\textsuperscript{\rm 3},
    Yuhang Guo\textsuperscript{\rm 1}\thanks{Corresponding authors.},
    Yangyang Kang\textsuperscript{\rm 2,3}\footnotemark[2]
}
\affiliations{
    \textsuperscript{\rm 1}Beijing Institute of Technology \ \ \ \
    \textsuperscript{\rm 2}Zhejiang University \ \ \ \
    \textsuperscript{\rm 3}ByteDance China \\
    \texttt{\{yaojiashu, hhy63, guoyuhang\}@bit.edu.cn, \ \ \{zengshuang, luochuwei, youwangjie, tangjie.jack, liuqingsong, yangyangkang\}@bytedance.com}
}

\begin{document}

\maketitle

\begin{abstract}
Through reinforcement learning (RL) with outcome correctness rewards, large reasoning models (LRMs) with scaled inference computation have demonstrated substantial success on complex reasoning tasks. However, the one-sided reward, focused solely on final correctness, limits its ability to provide detailed supervision over internal reasoning process. This deficiency leads to suboptimal internal reasoning quality, manifesting as issues like over-thinking, under-thinking, redundant-thinking, and disordered-thinking.
%Ultimately, this negatively impacts the final correctness of the outcome.
Inspired by the recent progress in LRM self-rewarding, we introduce self-rewriting framework, where a model rewrites its own reasoning texts, and subsequently learns from the rewritten reasoning to improve the internal thought process quality. For algorithm design, we propose a selective rewriting approach wherein only ``simple" samples, defined by the model's consistent correctness, are rewritten, thereby preserving all original reward signals of GRPO. For practical implementation, we compile rewriting and vanilla generation within one single batch, maintaining the scalability of the RL algorithm and introducing only $\sim 10\%$ overhead. Extensive experiments on diverse tasks with different model sizes validate the effectiveness of self-rewriting. In terms of the accuracy-length tradeoff, the self-rewriting approach achieves improved accuracy (+0.6) with substantially shorter reasoning (-46\%) even without explicit instructions in rewriting prompts to reduce reasoning length, outperforming existing strong baselines. In terms of internal reasoning quality, self-rewriting achieves significantly higher scores (+7.2) under the LLM-as-a-judge metric, successfully mitigating internal reasoning flaws.
\end{abstract}


\section{Introduction}

Using reinforcement learning (RL) for o1- and R1-like large reasoning model (LRM) post-training has demonstrated significant success in complex reasoning tasks with extended test-time computation \cite{shao2024deepseekmath, guo2025deepseek, qwq32b, wu2025inference}. The RL-based learning-to-reason paradigm provides trial and error reward signals focused on verifiable correctness \cite{xu2025towards}. This approach offers training flexibility and ideally holds the potential to achieve superhuman reasoning intelligence \cite{wan2024alphazero, hanposition, han2025general, silver2016mastering, silver2017mastering}.

\begin{figure}[htb!]
    \centering
    \includegraphics[width=1.0\linewidth]{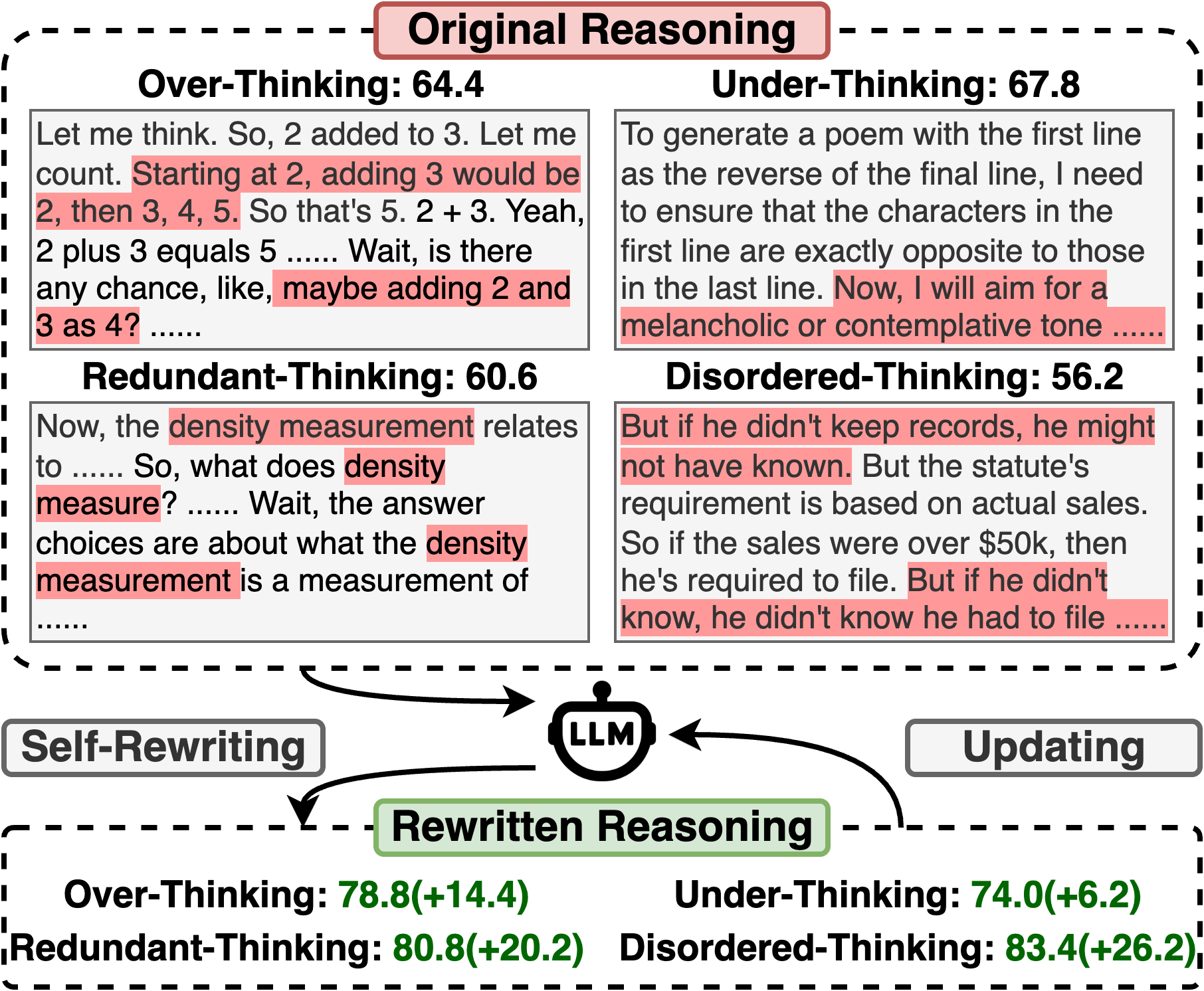}
    \caption{The reasoning of LRMs suffers from internal flaws. Over-thinking manifests as reasoning heavily over trivial or irrelevant aspects, under-thinking manifests as omissions and insufficient depth in reasoning, redundant-thinking manifests as repetition of essentially the same thoughts without bringing new ideas, and disordered-thinking manifests as jumping abruptly from one topic to another without coherence. Evaluated by LLM judges scores (the higher the better), rewriting significantly mitigates the flaws. Self-rewriting framework is to enable LRMs to learn from their own rewrites.}
    \label{fig:intro}
\end{figure}

However, a careful examination of reasoning texts generated by current LRMs reveals numerous internal flaws, as exemplified in Figure \ref{fig:intro}. Specifically, we identify four common problem categories in LRM reasoning: (1) over-thinking, where models heavily define, calculate, or check irrelevant or trivial parts; (2) under-thinking, characterized by skipping or over-simplifying complex and relevant parts; (3) redundant-thinking, involving the repetition of essentially the same thoughts without introducing new ideas; and (4) disordered-thinking, which manifests as the interleaving of multiple thinking threads into a confusing mess. Collectively, these problems compromise the interpretability of the result, incur unnecessary reasoning costs due to the generation of meaningless texts, ultimately degrade the correctness of the final outcomes \cite{su2025between}.

Although recent researches on post-tuning LRMs for reasoning improvement are emerging, these efforts primarily focus solely on one single aspect of length control \cite{aggarwal2025l1, yang2025towards, chen2024not}, neglecting the complex reasoning internal flaws.

To address the need for fine-grained reasoning improvement, we draw inspiration from recent research on self-rewarding in LRMs \cite{ryu2024multi, huang2023large, zhang2025right}. This self-supervised learning paradigm involves models providing their own internal rewards during the RL process. Instead of generating numerical rewards, we propose self-rewriting. This novel approach instructs models to rewrite their own generated reasoning passages to enhance quality while preserving all core ideas. The models then learn from these rewritten versions of their reasoning. As illustrated in Figure \ref{fig:intro}, preliminary experiments demonstrate that self-rewriting can significantly mitigate the four aforementioned problems, as measured by LLM-as-a-judge metrics.

Specifically, we integrate a rewriting process into GRPO \cite{shao2024deepseekmath} and define a corresponding rewriting preference reward. Our approach employs selective rewriting, focusing only on ``simple" queries where the model achieves 1.0 accuracy, while leaving other queries unchanged to minimize modifications to original GRPO algorithm. Furthermore, we designed an efficient implementation that introduces minimal computational overhead ($\sim10\%$) when incorporating the rewriting process and preference.

Through comprehensive evaluation across diverse tasks and LRMs of varying sizes, we've made several key findings. Firstly, rewriting effectively serves as a reasoning length control method, even without explicit instructions in the rewriting prompt to reduce length. Compared to existing methods designed for efficient reasoning, our approach yields comparable or superior results in the trade-off between length and accuracy, answering correctly with fewer reasoning tokens. Furthermore, a detailed analysis validates the effectiveness of rewriting in generating reasoning of varying length and mitigating internal reasoning flaws, including over-thinking, under-thinking, redundant-thinking, and disordered-thinking.

Our main contributions are summarized as follows:

\begin{itemize}
    \item We introduce the novel integration of reasoning rewriting into the LRM RL post-tuning framework to mitigate internal reasoning flaws. Our proposed self-rewriting framework enables models to learn from their rewritten reasoning, while maintaining the flexibility and scalability of the original GRPO algorithm.
    \item Extensive experiments across diverse tasks and multiple model sizes demonstrate that self-rewriting method exhibits strong length control capabilities, even without explicitly optimizing for length preference. It surpasses strong baselines in terms of the accuracy-length tradeoff.
    \item We conduct a fine-grained analysis of the rewritten texts. This validates that self-rewriting can generate more diverse response candidates of varying length, and successfully mitigate common reasoning flaws, including over-, under-, redundant-, and disordered-thinking. 
\end{itemize}

\section{Related Work}

\begin{figure*}[htb!]
    \centering
    \includegraphics[width=0.9\linewidth]{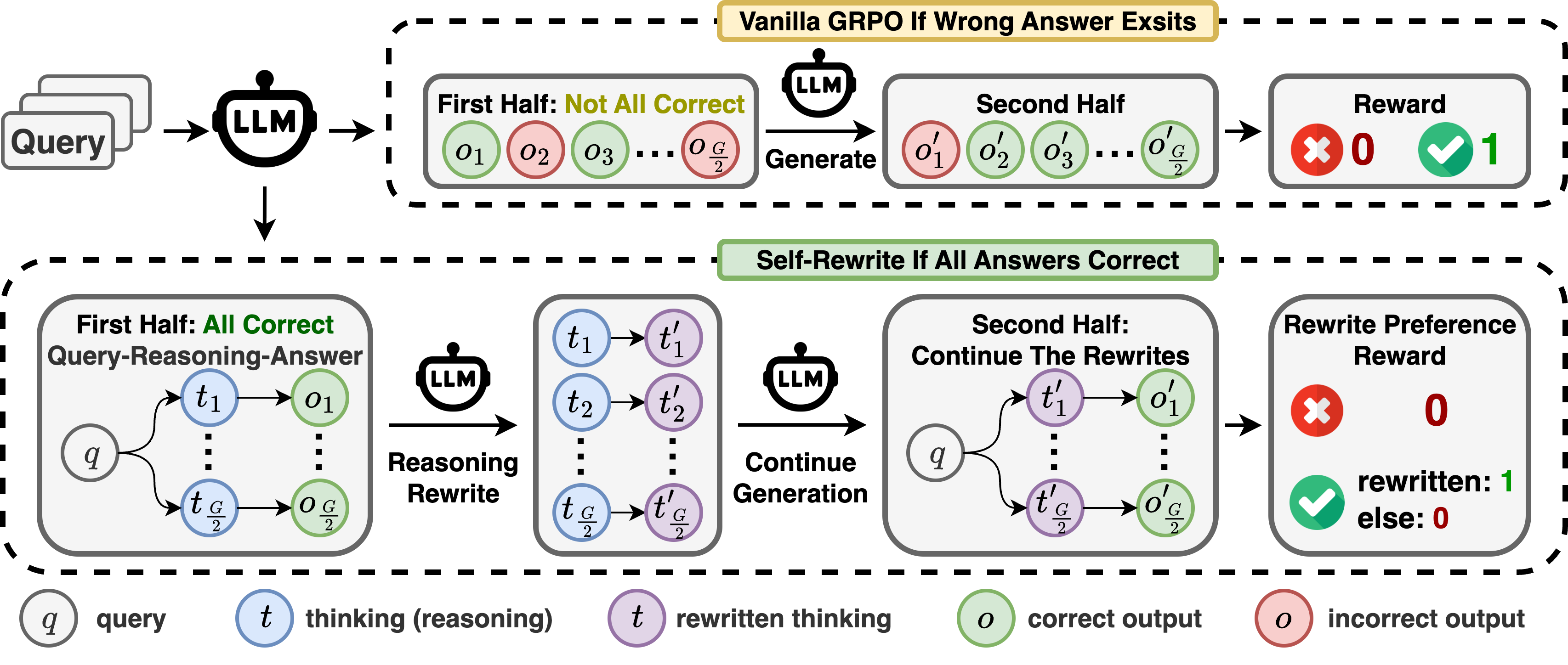}
    \caption{Self-rewriting framework. If any response for a given query is incorrect, the remaining half of the rollouts are sampled normally, and the final loss computation proceeds identically to GRPO. Conversely, if all initial responses for a query are correct, the model then rewrites and completes the reasoning texts, with the rewritten version receiving higher rewards.}
    \label{fig:method}
\end{figure*}

\paragraph{Reasoning Length Control}

Recent progress in LRMs \cite{guo2025deepseek, qwq32b, team2025kimi, seed2025seed1} has demonstrated the success of scaling test-time computation, where model performance on complex reasoning tasks consistently improves with the generation of more intermediate reasoning steps. However, a significant challenge for current LRMs is their tendency to generate an excessive number of reasoning tokens beyond necessary. This problem draws considerable research attention. One line of work explores modifying inference schemes to generate ``soft thoughts", where each represents an entire reasoning passage \cite{hao2024training, shen2025codi, zhang2025lightthinker}. Another approach endeavors to address the issue within the text modality itself. These works generate responses of varying lengths by: (1) sampling multiple times \cite{aggarwal2025l1, munkhbat2025self, su2025between}; (2) employing different prompts or thinking modes \cite{yang2025towards, sun2025empirical, zhang2025adaptthink}; or (3) explicitly truncating original responses into shorter ones \cite{chen2024not, qu2025optimizing, dai2025s}. The responses, now diverse in length, are then collected for fine-tuning, with a preference given to shorter and correct ones, ultimately aiming to generate concise and accurate reasoning paths.

Without explicit human-designed length preference, experiments show that self-rewriting method strikes a favorable accuracy-length tradeoff, outperforming strong baselines.

\paragraph{Reasoning Internal Quality Study}

Studying the internal quality of LRMs reasoning is more challenging than evaluating final accuracy or reasoning length, as the reasoning flaws cannot be directly formalized as an optimization objective. Nevertheless, reasoning quality, beyond just correctness and length, remains crucial for readability, interpretability, and overall performance \cite{su2025between}. Pioneering LRM research \cite{guo2025deepseek} leverages human-friendly cold-start fine-tuning to achieve readable formats and consistent language. More recently, \citet{qi2025models} proposes adapting reasoning to users' language to foster human trust and oversight. Another line of research \cite{lanham2023measuring, lyu2023faithful} investigates reasoning consistency and aims to enhance its influence on the final result. Recent related work \cite{wang2025thoughts} identifies a prevalent switching-thought pattern in LRM reasoning, and proposes intervening in the inference process to encouraging reasoning in depth.

Unlike previous work that heuristically selects one internal aspect for analysis—such as readability, consistency, or switching-thought frequency—we aim to comprehensively improve reasoning by leveraging self-rewriting.

\paragraph{LLM Self-Rewarding}

The LLM self-rewarding mechanisms as alternatives to human-labeled rewards in fine-tuning is becoming increasingly promising, particularly as the capabilities of LLMs approach or surpass human expert performance on numerous tasks. Some research advocates for the use of generative reward models \cite{ryu2024multi, mahan2024generative, yuan2024self, wu2024meta}, while others propose employing majority voting or leveraging internal representation relations to estimate ground truth labels \cite{huang2023large, zuo2025ttrl, zhang2025right}. The LLM self-rewarding paradigm not only decreases the intensive demand for human labeling, but also achieves performance comparable to direct training with ground-truth data.

Our work extend this line of work, proposing novel self-rewriting for self-supervised LLM finetuning. Comparing to self-rewarding, self-rewriting adopts an generative rather than discrimitive approach to self-improvement, offering a more detailed and comprehensive guide for the actor model.

\section{Method}
\label{sec:method}

Current LRM are supervised exclusively with verifable outcome rewards which focuses on the correctness of the results during RL training, lacking an explicit mechanism for internal quality improvement that comprehensively addressing reasoning flaws. These flaws compromise (1) the reasoning interpretability and oversight, (2) the reasoning efficiency, and (3) the final performance.

Inspired by recent advancements in self-rewarding mechanisms, we leverage the general language understanding and generation capabilities of LLMs to instruct models to rewrite their own reasoning passages. To enable learning from this rewritten reasoning while retaining the original performance, we integrate the selective rewriting process as a component into GRPO. Combining these elements, we propose the self-rewriting framework and present its efficient implementation.

\paragraph{Reasoning Rewriting}

Reasoning rewriting aims to change the features of the original reasoning and improve the overall quality. Addressing specific application scenarios, the rewriting instruction can be quite flexible, changing the reasoning style and so on. While in this work, we are to examine the basic principle of the method, thus only adopting a minimum rewriting instruction which focuses on general quality improvement and not involves special requirement. The general prompt for rewriting is shown in the text box below.

\begin{tcolorbox}[
    colback=gray!10,
    colframe=gray!75!black,
    fontupper=\footnotesize,
    boxsep=0pt
]
You are a skilled editor tasked with improving a given thinking passage. Your goal is to refine the passage to enhance its overall quality, making it more organized, coherent, and accurate. Your output should be a rewritten version of the original thinking passage. The rewritten version should maintain the core ideas and essence of the original while significantly improving its presentation and impact. Please provide only the rewritten thinking passage, without any additional explanations or context.
\end{tcolorbox}

\paragraph{Selective Rewriting}

By incorporating rewriting into the RL process, we aim to enable the LLM to learn from its own rewrites, thereby improving presentation and eliminating redundancy while retaining high reasoning effectiveness. Inspired by recent work that filters samples with 1.0 accuracy \cite{yu2025dapo}, we propose to apply selective rewriting, as illustrated in Figure \ref{fig:method}. This approach exclusively rewrites those samples that are already completely correct. Formally,
\begin{equation}
\label{eq:selective-rewriting}
\begin{aligned}
&\pi_\theta^{SR} (T, O | q, \{t_i, o_i\}_{i=1}^{\frac{G}{2}}) = \\
&\hspace{1cm}\begin{cases}
    \pi_\theta(T| \{t_i\}_{i=1}^{\frac{G}{2}}) \
    \pi_\theta(O| q, T) & \text{if } \{o_i\}_{i=1}^{\frac{G}{2}} \text{ all correct} \\
    \pi_\theta(T, O|q) & \text{else}
\end{cases}
\end{aligned}
\end{equation}
where $q, t, o$ denotes query, reasoning text, and final answer respectively. If and only if the first half group of responses are all correct, selective rewriting strategy rewrites the reasoning, consequently concatenates the original query and rewritten reasoning for continue generation of final answers. Otherwise, the following half group are sampled normally.

Selective rewriting strategy offers two key advantages. Firstly, it minimally interrupts current optimization methods, utilizing only what would otherwise be useless samples in the original GRPO process. Secondly, by focusing on the simplest queries that the LLM has already mastered, it allows the model to practice generating concise and well-presented thoughts on easy problems, while still engaging in normal sampling for harder ones, preserving the improvement space for complex problem-solving.

\paragraph{Optimization}

Self-rewriting framework gives higher rewards to rewritten samples if all responses are correct; otherwise, it gives higher rewards to correct samples. Formally, given the raw correctness reward $\mathbf{r}$, the reward for the $i$-th sample in a group is defined as

\begin{equation}
\label{eq:ir}
\begin{aligned}
    \hat{r}_i =
    \begin{cases}
        r_i & \text{if  } \mathbf{r} \neq \textbf{1}, \\
        1 & \text{if  } \mathbf{r} = \textbf{1} \land (i\text{-th sample is rewritten}), \\
        0 & \text{else.}
    \end{cases}
\end{aligned}
\end{equation}

Calculate the advantage $\hat{A}_{i,t}$ based on $\hat{\mathbf{r}}$ (details are discussed in Appendix \ref{app:advantage}), the policy model is optimized by maximizing the objective:
\begin{equation}
\label{eq:grpo-loss}
\begin{aligned}
&\mathcal{J}_\theta = \mathbb{E} \Big[
q \sim P(Q),
\{t, o\}_{i=1}^{\frac{G}{2}} \sim \pi_{\theta_{old}}(\cdot|q), \\
& \quad \quad \quad \ \  \{t, o\}_{i=\frac{G}{2}+1}^{G} \sim \pi_{\theta_{old}}^{SR} (\cdot | q, \{t_i, o_i\}_{i=1}^{\frac{G}{2}}) \Big] \\
&\frac{\sum_{i=1}^{G}}{G}  \frac{\sum_{t=1}^{|o_i|}}{|o_i|}  
\Big\{ \min [ \frac{\pi_\theta}{\pi_{\theta_{\text{old}}}}\hat{A}_{i,t},
\text{clip} ( \frac{\pi_\theta}{\pi_{\theta_{\text{old}}}}, \, 1 \pm \epsilon )\hat{A}_{i,t}] \Big\}.
\end{aligned}
\end{equation}

\paragraph{Overall Algorithm}
The pseudocode of self-rewriting is shown in Algorithm \ref{alg:ours}.
\begin{algorithm}[htbp]
\caption{\textsc{Self-Rewriting}}
\label{alg:ours}
\textbf{Input}: query set $Q$, model $\mathcal{M}$, verifier $\mathcal{R}$, group size $G$ \\
\textbf{Output}: adapted model $\mathcal{M}$
\begin{algorithmic}[1] %[1] enables line numbers
\FOR{$q$ in $Q$}

\STATE $\mathbf{t}_{1:\frac{G}{2}}, \mathbf{o}_{1:\frac{G}{2}} \gets \mathcal{M}(q)$
\algorithmiccomment{generate $\frac{G}{2}$ responses}
\STATE $\mathbf{r}_{1:\frac{G}{2}} \gets \mathcal{R}(q, \mathbf{o}_{1:\frac{G}{2}})$
\algorithmiccomment{verify first half batch}

\IF{$\mathbf{r}_{1:\frac{G}{2}} = \mathbf{1}$}
    \STATE $\mathbf{t}_{\frac{G}{2}+1:G} \gets \mathcal{M}(\mathbf{t}_{1:\frac{G}{2}})$
    \algorithmiccomment{rewrite reasoning}
    \STATE $\mathbf{o}_{\frac{G}{2}+1:G} \gets \mathcal{M}(q, \mathbf{t}_{\frac{G}{2}+1:G})$
    \algorithmiccomment{continue generation}
\ELSE
    \STATE $\mathbf{t}_{\frac{G}{2}+1:G}, \mathbf{o}_{\frac{G}{2}+1:G} \gets \mathcal{M}(q)$
    \algorithmiccomment{vanilla generation}
\ENDIF

\STATE $\mathbf{r}_{1:G} \gets \mathcal{R}(q, \mathbf{o}_{1:G})$
\STATE $\mathbf{\hat{r}}_{1:G} \gets \text{Equ2}(\mathbf{r}_{1:G})$
\algorithmiccomment{reward with Equation \ref{eq:ir}}
\STATE $\mathcal{M} \gets step(\mathcal{M}, q, \mathbf{t}, \mathbf{o}, \mathbf{\hat{r}})$
\algorithmiccomment{GRPO stepping}

\ENDFOR
\STATE \textbf{return} $\mathcal{M}$
\end{algorithmic}
\end{algorithm}

\paragraph{Efficient Implementation}

\begin{figure}[htb!]
    \centering
    \includegraphics[width=0.95\linewidth]{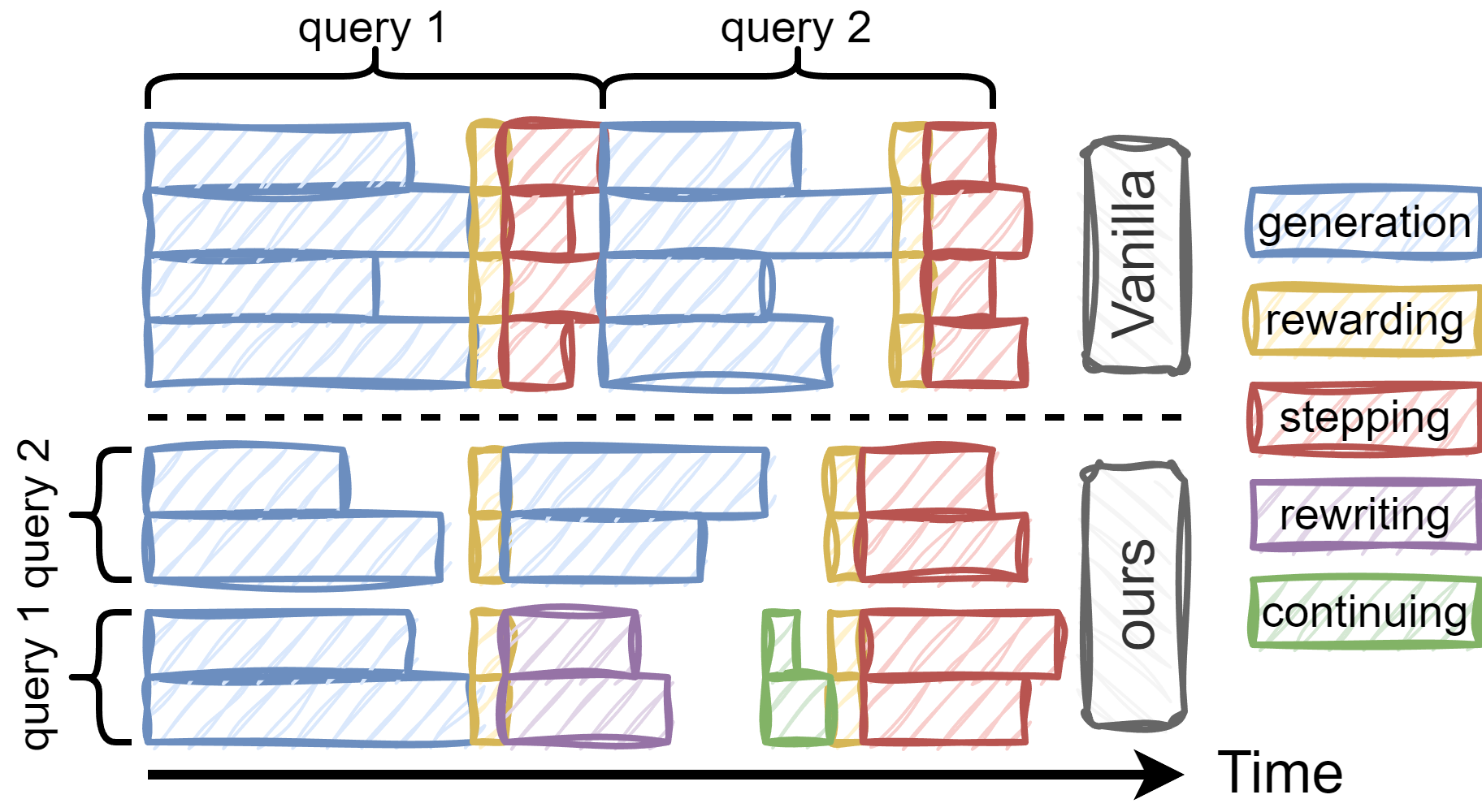}
    \caption{The illustration of computation process for vanilla GRPO (top) and self-rewriting (bottom), when there are $4$ computation units, $2$ queries in a batch, and $4$ rollouts for each query. By compiling generation and rewriting into one batch, the complexity introduced by selective rewriting only leads to time consumption increment of about 10\%.}
    \label{fig:efficient}
\end{figure}

Self-rewriting complicates the RL process by incorporating a selective rewriting after generating the first half batch. To ensure the added complexity does not lead to significantly higher computation consumption, we design a practically efficient implementation for the algorithm. As is shown in Figure \ref{fig:efficient}, after the generation of the first half batch, we compile the vanilla generation samples from not-all-correct queries and the rewriting samples from all-correct queries into one batch for joint inference. Subsequently, the rewritten reasoning segments are separately compiled into another batch for continued generation. Note that the continuation only generates outputs after \texttt{</think>} token, which usually takes a very small portion of the overall response. Compared to vanilla GRPO, the implementation increases the time consumption by only about 10\%, which is acceptable given that our method is primarily intended for post-tuning rather than large-scale pre-training.

\section{Experiments}

\begin{table*}[htb!]
    \centering
    \setlength{\tabcolsep}{1mm}
    \begin{tabular}{lcccccccccccc|ccc}
        \toprule
        \multirow{2}{*}{\textbf{Models}} & \multicolumn{3}{c}{\textbf{MATH-500}} & \multicolumn{3}{c}{\textbf{GPQA-Diamond}} & \multicolumn{3}{c}{\textbf{ARC-Challenge}} & \multicolumn{3}{c}{\textbf{MMLU-Pro}} & \multicolumn{3}{|c}{\textbf{Average}} \\
        \cmidrule(lr){2 - 4} \cmidrule(lr){5 - 7} \cmidrule(lr){8 - 10} \cmidrule(lr){11 - 13} \cmidrule(lr){14 - 16} & \textbf{Acc} & \textbf{Len} & \textbf{Jdg} & \textbf{Acc} & \textbf{Len} & \textbf{Jdg} & \textbf{Acc} & \textbf{Len} & \textbf{Jdg} & \textbf{Acc} & \textbf{Len} & \textbf{Jdg} & \textbf{Acc} & \textbf{Len} & \textbf{Jdg}\\
        \midrule
        \rowcolor{gray!20} \multicolumn{16}{c}{\it Qwen3-1.7B} \\
        Original &
        86.0 & 5267 & 71.5 & 
        37.9 & 7999 & 51.0 &
        87.0 & 798 & 74.6 &
        56.6 & 3354 & 64.3 &
        66.9(+0.0) & 4355(+0\%) & 65.4(+0.0) \\
        GRPO &
        87.0 & 4695 & 71.2 &
        37.9 & 6960 & 50.7 &
        86.1 & 717 & 77.1 &
        56.6 & 3105 & 66.6 &
        \bf 66.9(+0.0) & 3869(-11\%) & 66.4(+1.0) \\
        LenPen1 &
        83.8 & 3794 & 77.5 &
        33.8 & 5637 & 52.5 &
        87.3 & 591 & 81.4 &
        56.4 & 2381 & 69.4 &
        65.3(-1.6) & 3101(-29\%) & 70.2(+4.8) \\
        LenPen2 &
        83.7 & 3893 & 74.4 &
        34.2 & 5814 & 52.7 &
        87.1 & 597 & 80.8 &
        55.9 & 2527 & 68.6 &
        65.2(-1.7) & 3209(-27\%) & 69.1(+3.7) \\
        ShortBetter & 
        86.0 & 3714 & 75.4 &
        37.1 & 5947 & 53.0 &
        86.1 & 607 & 82.1 &
        56.4 & 2577 & 68.1 &
        66.4(-0.5) & 3211(-26\%) & 69.7(+4.3) \\
        LPO &
        83.4 & 3854 & 73.0 &
        36.0 & 4974 & 53.1 &
        86.0 & 641 & 80.6 &
        55.5 & 2386 & 65.2 &
        65.2(-1.7) & 2964(-32\%) & 68.0(+2.6)\\
        TOPS &
        80.7 & 3855 & 74.3 &
        36.2 & 4632 & 51.4 &
        86.3 & 580 & 80.9 &
        55.3 & 2238 & 68.1 &
        64.6(-2.3) & \bf 2826(-35\%) & 68.7(+3.3) \\
        % Rewrite-off  & \\
        Rewrite &
        86.0 & 3820 & 79.6 &
        36.9 & 4823 & 55.5 &
        86.3 & 503 & 84.9 & 
        55.7 & 2195 & 74.0 &
        66.2(-0.7) & 2835(-35\%) & \bf 73.5(+8.1) \\
        \midrule

        \rowcolor{gray!20} \multicolumn{16}{c}{\it Qwen3-4B} \\
        Original &
        89.6 & 4544 & 73.4 &
        52.2 & 7373 & 54.5 &
        92.7 & 742 & 75.4 &
        69.9 & 3440 & 64.9 &
        76.1(+0.0) & 4025(+0\%) & 67.1(+0.0) \\
        GRPO &
        90.0 & 3910 & 74.7 &
        52.5 & 7057 & 54.3 &
        92.4 & 692 & 77.7 &
        68.1 & 3063 & 67.8 &
        75.8(-0.3) & 3681(-9\%) & 68.6(+1.5) \\
        LenPen1 &
        89.2 & 2634 & 81.4 &
        54.0 & 4650 & 57.8 &
        92.5 & 480 & 83.2 &
        69.4 & 2041 & 73.2 &
        76.3(+0.2) & 2451(-39\%) & 73.9(+6.8)\\
        LenPen2 &
        89.4 & 2607 & 81.9 &
        52.0 & 4966 & 57.8 &
        92.9 & 491 & 83.0 &
        69.6 & 2091 & 70.3 &
        76.0(-0.1) & 2539(-37\%) & 73.3(+6.2) \\
        ShortBetter &
        89.2 & 2480 & 84.6 &
        49.2 & 4904 & 57.3 &
        92.8 & 499 & 82.4 &
        67.8 & 2196 & 72.6 &
        74.8(-1.3) & 2520(-37\%) & 74.2(+7.1) \\
        LPO &
        89.4 & 3142 & 79.6 &
        48.8 & 4602 & 57.3 &
        92.1 & 598 & 79.4 &
        66.8 & 2277 & 69.0 &
        74.3(-1.8) & 2655(-34\%) & 71.3(+4.2) \\
        TOPS &
        88.6 & 2929 & 85.2 &
        46.8 & 3772 & 55.3 &
        92.2 & 419 & 81.0 &
        66.6 & 1694 & 73.0 &
        73.6(-2.5) & \bf 2204(-45\%) & 73.6(+6.5)
        \\
        % Rewrite-off  & \\
        Rewrite &
        89.4 & 2005 & 88.8 &
        54.6 & 4647 & 58.6 &
        93.0 & 468 & 85.8 &
        68.9 & 1747 & 76.2 &
        \bf 76.5(+0.4) & 2217(-45\%) & \bf 77.4(+10.3) \\
        \midrule

        \rowcolor{gray!20} \multicolumn{16}{c}{\it Qwen3-8B} \\
        Original &
        90.2 & 4663 & 78.6 &
        55.1 & 8063 & 58.5 &
        93.5 & 764 & 82.1 &
        74.1 & 3482 & 69.3 &
        78.2(+0.0) & 4243(+0\%) & 72.1(+0.0) \\
        GRPO &
        89.6 & 4446 & 78.6 &
        53.0 & 7178 & 56.9 &
        93.9 & 728 & 82.5 &
        76.4 & 3194 & 69.8 &
        78.2(+0.0) & 3887(-8\%) & 72.0(-0.1) \\ 
        LenPen1 &
        89.0 & 3034 & 83.6 &
        55.6 & 5371 & 58.8 &
        93.9 & 548 & 85.4 &
        74.3 & 2353 & 72.1 &
        78.2(+0.0) & 2827(-33\%) & 75.0(+2.9) \\
        LenPen2 &
        89.2 & 2938 & 82.9 &
        56.6 & 5929 & 59.0 &
        94.0 & 560 & 84.5 &
        74.3 & 2380 & 73.5 &
        78.5(+0.3) & 2952(-30\%) & 75.0(+2.9) \\
        ShortBetter &
        90.4 & 3001 & 83.7 &
        53.3 & 5754 & 58.0 &
        93.7 & 560 & 86.6 &
        74.5 & 2402 & 72.9 &
        78.0(-0.2) & 2929(-31\%) & 75.3(+3.2) \\
        LPO &
        89.6 & 3648 & 80.9 &
        52.3 & 5666 & 56.7 &
        94.1 & 625 & 82.3 &
        73.8 & 2698 & 70.6 &
        77.5(-0.7) & 3159(-26\%) & 72.6(+0.5) \\
        TOPS &
        89.3 & 2796 & 83.4 &
        53.8 & 4505 & 56.7 &
        94.2 & 474 & 86.4 &
        73.9 & 2048 & 72.4 &
        77.8(-0.4) & 2456(-42\%) & 74.7(+2.6) \\
        % Rewrite-off  & \\
        Rewrite &
        89.2 & 2490 & 90.0 &
        57.1 & 4018 & 61.2 &
        94.3 & 463 & 89.5 &
        74.4 & 2202 & 76.6 &
        \bf 78.8(+0.6) & \bf 2293(-46\%) & \bf 79.3(+7.2) \\
        \bottomrule
    \end{tabular}
    \caption{Main experimental results. Best average results for each model are bolded.}
    \label{tab:main-table}
\end{table*}

\subsection{Setups}

We list main experimental setups here, and detailed experimental information are shown in Appendix \ref{app:prompts} and \ref{app:implementation}.

\paragraph{Datasets}
For training datasets, we randomly sample 10K pieces from DeepMath-103K \cite{deepmath} following previous work \cite{dai2025s}. For comprehensive evaluation, we conduct test on four diverse types of tasks, including math reasoning MATH-500 \cite{lightman2023let}, science reasoning GPQA-Diamond \cite{rein2024gpqa}, logic reasoning ARC-Challenging \cite{clark2018think}, and knowledge reasoning MMLU-Pro \cite{wang2024mmlu}.

\paragraph{Metrics}
Following previous work, we firstly measure the tradeoff between accuracy and length. Specifically, we evaluate the average $pass@1$ rate and token numbers across $4$ sampled runs with temperature $0.6$ and max length 32K. In addition, we are also interested in the internal quality of reasoning texts, i.e., the over-thinking, under-thinking, redundant-thinking, and disordered-thinking problems. However, LLM reasoning texts are often very long and require professional level knowledge to understand, thus very difficult to evaluate. For efficient and effective evaluation, we use stronger LLMs (DeepSeek-V3, others in Appendix \ref{app:other-judges}) instructed to judge the reasoning text according to the four aspects (prompts shown in Appendix \ref{app:prompts}), and score in range of $1$ to $5$. We scale the averaged results into $100$.

\paragraph{Models}
We conduct experiments with well-established LRMs with varied parameter sizes, including Qwen3-1.7B, Qwen3-4B, and Qwen3-8B.

\paragraph{Baselines}
(1) GRPO \cite{shao2024deepseekmath} reinforces correctness rewards. (2) Length Penalty is a simple but effectively method widely used by previous work \cite{team2025kimi, hiroshi2025practical, arora2025training}, which heuristically adopts length rewards as supplements to correctness rewards favoring shorter responses. We implements two types of length penalty proposed by \citet{team2025kimi} and \citet{arora2025training}, referring them as LP1 and LP2 respectively. (3) ShorterBetter \cite{yi2025shorterbetter} prefers responses whose length close to the shortest correct ones, instead of simply shorter ones. (4) LPO \cite{su2025between} leverage offline-RL to prefer shorter responses over longer ones. (5) TOPS \cite{yang2025towards} applies reasoning effort-conditioned generation for multiple responses with diverse conciseness level, and adopts offline-RL preferring the shortest correct reasoning.

\paragraph{Implementation}
For result reliability and robustness, we adopt commonly-used hyper-parameters without extensive search. All methods are trained for $1$ epoch with batch size $256$. For online RL, we use GRPO with learning rate as $3e-6$ and rollout size as $8$. For offline RL, we use SimPO \cite{meng2024simpo} with learning rate as $1e-6$, $\beta = 2.0$ and $\gamma = 0.3$. Other details are listed in Appendix \ref{app:implementation}.

\subsection{Results}

The main results are shown in Table \ref{tab:main-table}.

\paragraph{Accuracy-Length Tradeoff}
All methods experimented manage to shorten reasoning length, including vanilla GRPO with a cutoff length 12K. Among them, our proposed self-rewriting method strikes a good tradeoff between accuracy and reasoning length, outperforming existing strong baselines. Specifically, with in the same training load, self-rewriting is able to generate more concise reasoning leading to more correct final answers, due to the strength of LLM rewriting supervision over previous length preference.

\paragraph{LLM Judge Scores}
LLM judge scores assess issues such as over-thinking, under-thinking, redundant-thinking, and disordered-thinking in reasoning texts by leveraging the general language understanding capabilities of powerful LLMs. The results reveal that stronger models yield higher judge scores, while length control methods can further enhance the internal quality of reasoning. Notably, self-rewriting achieves significantly higher LLM judge scores compared to all baseline methods, demonstrating its capacity to improve internal reasoning quality and alleviate reasoning flaws. These substantially higher judge scores partially explain why self-rewriting achieves a favorable accuracy-length tradeoff.

\paragraph{Online versus Offline}

Our comparison between online RL methods (LenPen1, LenPen2, ShorterBetter, and our proposed self-rewriting) and offline RL methods (LPO and TOPS) reveals two key distinctions. First, regarding final accuracy and LLM judge scores, online methods generally outperform offline approaches even when offline methods like TOPS also incorporate correctness preferences. This demonstrates the importance of online generation for maintaining performance, as models require dynamic reward signal calibration alongside parameter updates, rather than relying on static preference datasets constructed before fine-tuning begins. Second, online RL methods demonstrate more proportional length control than offline approaches on out-of-domain tasks. When trained on math reasoning data, offline methods tend to over-truncate GPQA reasoning which requires more detailed explanations, while under-truncating simpler ARC tasks. In contrast, online methods adjust lengths more appropriately for all tasks. In summary, for length control scenarios requiring both accuracy preservation and generalization beyond training data, online methods, particularly our proposed self-rewriting, offer superior performance.

\subsection{Ablations}

We conduct further ablation experiments to answer two key questions. (1) Does our proposed self-rewriting method consistently outperform its baselines across different length compression scales? (2) How does selective rewriting policy influences the final results?

\paragraph{Training Data Scale}

\begin{figure}[htb!]
    \centering
    \includegraphics[width=1.0\linewidth]{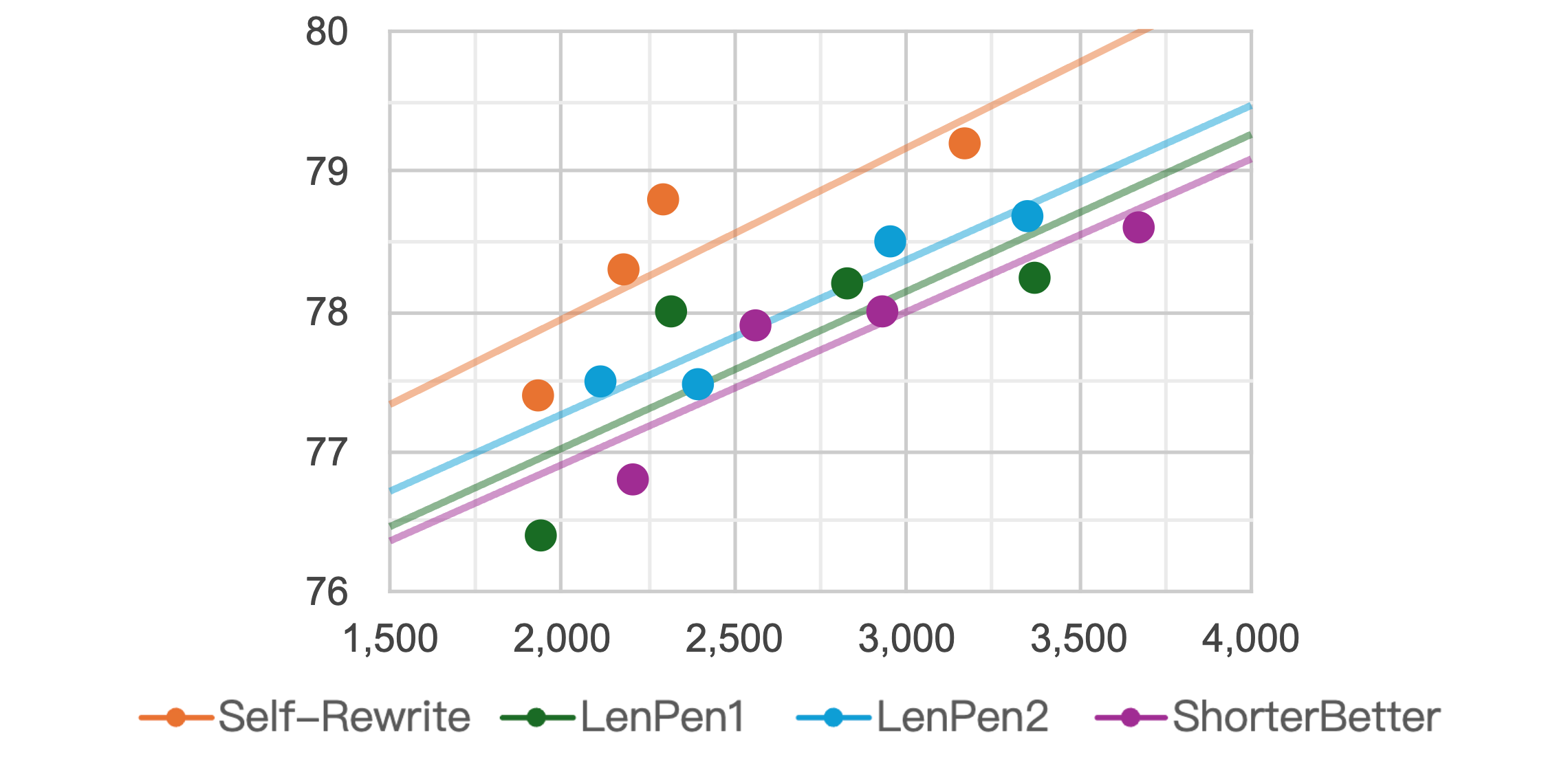}
    \caption{Results of four checkpoints for each methods on Qwen3-8B. X-axis refers to length and y-axis refers to average accuracy across four tasks.}
    \label{fig:scatter}
\end{figure}

To compare our self-rewriting method with its baselines across different lengths, we continue training the online RL methods for an additional 10K samples (20K samples in total). We then evaluate four checkpoints corresponding to 5K, 10K, 15K, and 20K training samples for each method. The results are presented in Figure \ref{fig:scatter}.

The results show that, with more training, length control methods progressively shorten reasoning length at the cost of compromising accuracy. Among them, self-rewriting consistently outperforms other online RL baselines for accuracy under different length budgets.

\paragraph{Selective Rewriting}

We propose selective rewriting to let the model exclusively rewrite ``simple" samples for better presentation, while explore for correctness for other ones. To validate the effectiveness of selective rewriting, we compare it with a vanilla rewriting policy, where at each step, $n\%$ of samples are randomly chosen for rewriting. As the results in Table \ref{tab:ablation} show, the performance is degraded without selective rewriting process.

\begin{table}[htb!]
\setlength{\tabcolsep}{1.5mm}
    \centering
    \begin{tabular}{lccc}
        \toprule
        \textbf{Models} & \textbf{Acc} & \textbf{Len} & \textbf{Jdg}\\
        \midrule

        Original &
        78.2(+0.0) & 4243(+0\%) & 72.1(+0.0) \\

        GRPO &
        78.2(+0.0) & 3887(-8\%) & 72.0(-0.1) \\ 

        Self-Rewriting &
        \bf{78.8(+0.6)} & 2293(-46\%) & \bf{79.3(+7.2)} \\

        \quad w/o SR($50\%$) & 77.9(-0.3) & 2691(-37\%) & 78.8(+6.7) \\
        \quad w/o SR($100\%$) & 77.8(-0.4) & \bf{2120(-50\%)} & 78.6(+6.5) \\
        \bottomrule
    \end{tabular}
    \caption{Ablation experimental results on Qwen3-8B. All results are averaged on four tasks. ``SR" denotes selective rewriting, and ``without SR ($n$\%)" refers to the process that randomly select $n\%$ samples to rewrite, instead of choosing correct ones as in selective rewriting. Best results are bolded.}
    \label{tab:ablation}
\end{table}

\section{Analysis}

In previous experiments, we have validated the high effectiveness of our proposed self-rewriting over previous length control methods in terms of accuracy-length tradeoff, and we attribute such superiority to the high reasoning text understanding and rewriting capability of LLMs. Specifically, we speculate that LLMs are able to rewrite the reasoning texts generated by their own for higher internal reasoning quality, mitigating the issues including over-thinking, under-thinking, redundant-thinking, and disordered-thinking.

While our hypothesis that self-rewriting can improve reasoning internal quality has been validated through the LLM-as-a-judge methods of previous experiments as shown in Table \ref{tab:main-table}, we want to have a deeper understanding about how rewriting changes the character of reasoning texts in details. In this section, we conduct further analyses to offer more insights about how rewriting improve the reasoning texts.

Before presenting detailed results, we emphasize that all rewriting analyses are based on the neutral prompts described in Section \ref{sec:method}. Our tested rewriting prompt uses only basic instructions to improve reasoning quality generically, without revealing our specific evaluation criteria (over-thinking, etc.) or explicitly requesting conciseness. While application-specific scenarios could employ more targeted prompts, we intentionally use minimal instructions to enable fundamental evaluation of the core methodology of self-rewriting.

\subsection{Length Ratio Distribution}

Previous reasoning length control approaches typically incorporate length preference as supplements to correctness rewards, while self-rewriting framework rewrites reasoning texts and prefers the rewritten responses in rewarding. We first examine the distribution of length ratios between preferred and rejected responses under self-rewriting approach and previous length preference approaches.

In practice, we analyze the first $20$ steps (corresponding to 5K samples) with Qwen3-8B self-rewriting training, and count the ratio between preferred responses (rewritten ones) and rejected responses (original ones). We compare the length ratio distribution with that of previous length control approaches, which denotes the ratio between the shortest correct responses and the other correct responses.

\begin{figure}[htb!]
    \centering
    \includegraphics[width=1.0\linewidth]{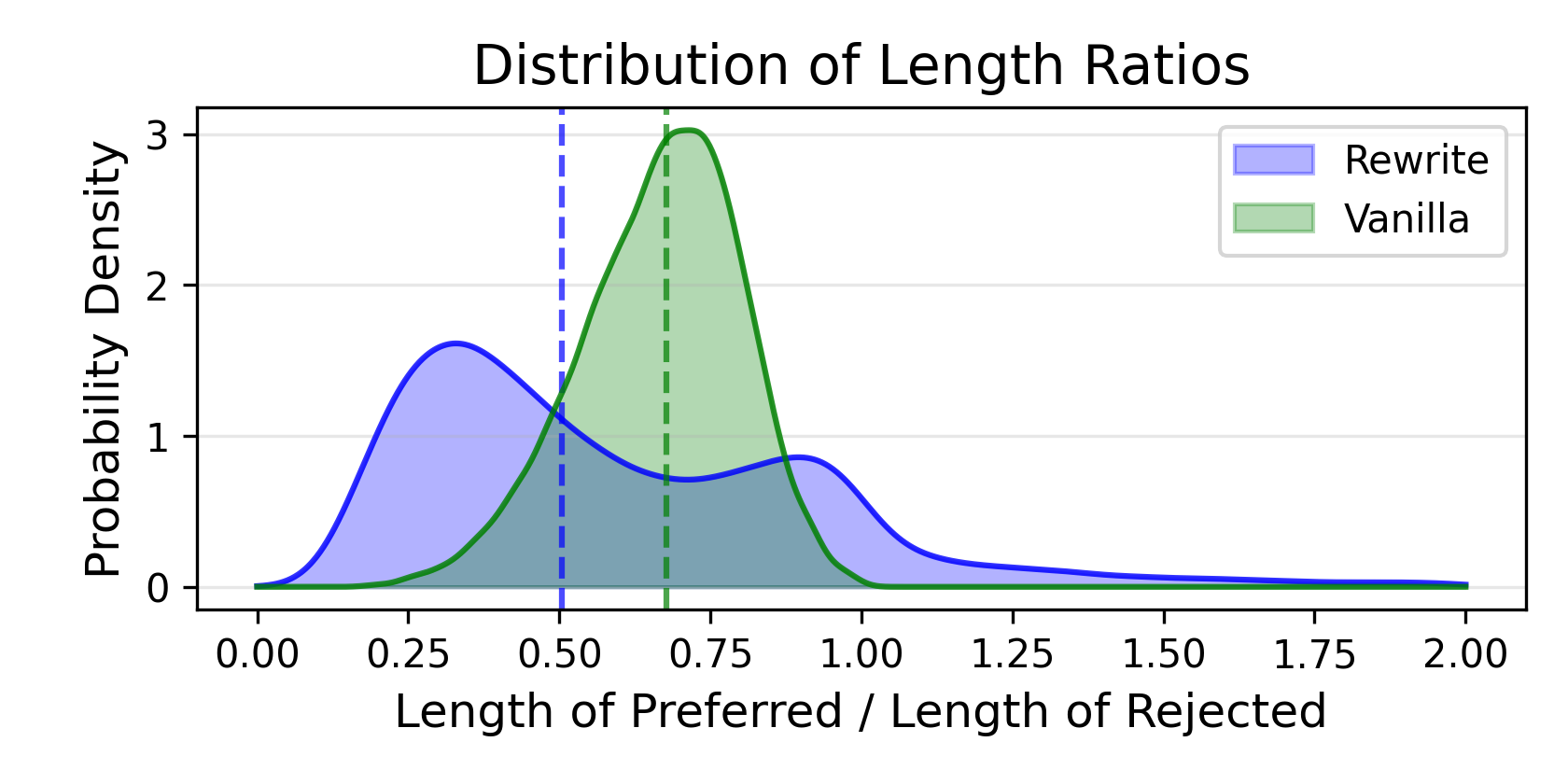}
    \caption{The length ratio distribution of Qwen3-8B online RL training dataset between preferred and other responses. Preferred responses refer to rewritten ones for self-rewriting, and shortest correct ones for vanilla length control approaches. The vertical dashed lines denote median values. The probability density function shown in the figure is obtained using Gaussian kernel density estimation.}
    \label{fig:ratio}
\end{figure}

The results in Figure \ref{fig:ratio} reveal three key insights. First, self-rewriting in general adopts a more aggressive length reduction strategy, with a median length ratio of $\sim$0.5 compared to $\sim$0.7 for vanilla sampling methods. This explains Table \ref{tab:main-table}'s findings where self-rewriting achieves shorter responses with equivalent training. Second, self-rewriting exhibits significantly greater variance in length ratios, indicating its greater ability to generate more diverse response candidates in terms of length. Notably, $\sim$10\% of rewritten samples become longer than originals, which is an outcome impossible with vanilla length preferences. Finally, self-rewriting displays a bimodal distribution, rather than the unimodal distribution pattern of vanilla length preference, suggesting LLMs dynamically adapt conciseness based on problem characteristics rather than applying uniform length reduction.

\subsection{Fine-Grained LLM Judge Scores}

Our discussion has revealed the internal reasoning flaws in current LRMs as shown in Figure \ref{fig:intro}, manifesting as over-thinking, under-thinking, redundant-thinking, and disordered-thinking. Moreover, through rigorous LLM-as-a-judge evaluation with powerful model evaluators, we have demonstrated the self-rewriting's significant mitigation of these issues, as shown in Table \ref{tab:main-table}. We now conduct a fine-grained examination of how rewriting enhances reasoning quality.

Using the same judge model and prompts from our main experiments, we analyze fine-grained scores for each reasoning flaw category: over-thinking, under-thinking, redundant-thinking, and disordered-thinking. Our primary evaluation focuses on the DeepMath training data, which is also used in our main experiments, to understand how self-rewriting improves reasoning quality. For generalizability, we conduct additional testing on the MMLU training set. 1K queries are sample from both datasets.

\begin{table}[htb!]
\setlength{\tabcolsep}{1.7mm}
    \centering
    \begin{tabular}{lcccc}
        \toprule
        \textbf{Models} & \textbf{Over} & \textbf{Under} & \textbf{Redundant} & \textbf{Disordered}\\
        \midrule

        \rowcolor{gray!20} \multicolumn{5}{c}{\it DeepMath} \\
        Original & 76.3 & 82.0 & 67.9 & 61.6 \\
        Rewritten & 84.0 & 87.3 & 82.1 & 77.1 \\

        \rowcolor{gray!20} \multicolumn{5}{c}{\it MMLU} \\
        Original & 64.4 & 67.8 & 60.6 & 56.2 \\
        Rewritten & 78.8 & 74.0 & 80.8 & 83.4 \\
        
        \bottomrule
    \end{tabular}
    \caption{LLM-as-a-judge scores of reasoning texts on four dimensions, including over-thinking, under-thinking, redundant-thinking, and disordered-thinking. The original the rewritten reasoning texts are all generated by Qwen3-8B.}
    \label{tab:analysis}
\end{table}

The results in Table \ref{tab:analysis} reveal consistent patterns across both datasets. Redundant-thinking and disordered-thinking show significant improvement evidenced by higher scores, while over-thinking and under-thinking demonstrate more modest gains. This aligns with expectations, as LLM reflection can more easily identify repetitive content and incoherent topic transitions compared to determining optimal thought depth and coverage. Notably, direct rewriting yields greater LLM judge score improvements than fine-tuning with our self-rewriting framework. This suggests the online RL framework's correctness scores prevent overfitting to rewritten outputs, maintaining (or even improving) accuracy despite smaller judge score gains, demonstrating an effective balance.

\section{Conclusion}
This paper proposes self-rewriting, an RL reasoning reinforcement framework that incorporated with model-generated rewrites. Self-rewriting framework employs selective rewriting to minimize disruption to the vanilla RL process, focusing exclusively on simple queries while favoring their rewritten reasoning texts. Our implementation optimizes efficiency by compiling continuous generation and rewriting into a single batch operation, obtaining a computational overhead as low as $\sim10\%$. Extensive experiments across diverse tasks and model sizes validate the superior reasoning improvement ability of self-rewriting over strong baselines, particularly in balancing accuracy and length while achieving higher reasoning quality according to LLM judge metrics.

Self-rewriting extends current line of self-rewarding approaches for reasoning improvement, demonstrating that LLMs can generate concise high-quality supervision through rewriting their own reasoning texts. The framework's flexibility allows for targeted rewriting by modifying prompt instructions, enabling the generation of texts with specific features tailored to particular applications. Tthis work focuses on general rewriting instructions, and leaves the exploration of specialized targeted rewriting for future research.

\section*{Acknowledgements}
This work is supported by the National Natural Science Foundation of China (Grant No. U21B2009).

\bibliography{aaai2026}

@inproceedings{wu2025inference,
  title={Inference scaling laws: An empirical analysis of compute-optimal inference for LLM problem-solving},
  author={Wu, Yangzhen and Sun, Zhiqing and Li, Shanda and Welleck, Sean and Yang, Yiming},
  booktitle={The Thirteenth International Conference on Learning Representations},
  year={2025}
}

@article{shao2024deepseekmath,
  title={Deepseekmath: Pushing the limits of mathematical reasoning in open language models},
  author={Shao, Zhihong and Wang, Peiyi and Zhu, Qihao and Xu, Runxin and Song, Junxiao and Bi, Xiao and Zhang, Haowei and Zhang, Mingchuan and Li, YK and others},
  journal={arXiv preprint arXiv:2402.03300},
  year={2024}
}

@article{guo2025deepseek,
  title={Deepseek-r1: Incentivizing reasoning capability in llms via reinforcement learning},
  author={Guo, Daya and Yang, Dejian and Zhang, Haowei and Song, Junxiao and Zhang, Ruoyu and Xu, Runxin and Zhu, Qihao and Ma, Shirong and Wang, Peiyi and Bi, Xiao and others},
  journal={arXiv preprint arXiv:2501.12948},
  year={2025}
}

@misc{qwq32b,
    title = {QwQ-32B: Embracing the Power of Reinforcement Learning},
    url = {https://qwenlm.github.io/blog/qwq-32b/},
    author = {Qwen-Team},
    month = {March},
    year = {2025}
}

@article{team2025kimi,
  title={Kimi k1. 5: Scaling reinforcement learning with llms},
  author={Kimi-Team and Du, Angang and Gao, Bofei and Xing, Bowei and Jiang, Changjiu and Chen, Cheng and Li, Cheng and Xiao, Chenjun and Du, Chenzhuang and Liao, Chonghua and others},
  journal={arXiv preprint arXiv:2501.12599},
  year={2025}
}

@article{seed2025seed1,
  title={Seed1. 5-thinking: Advancing superb reasoning models with reinforcement learning},
  author={Seed, ByteDance and Chen, Jiaze and Fan, Tiantian and Liu, Xin and Liu, Lingjun and Lin, Zhiqi and Wang, Mingxuan and Wang, Chengyi and Wei, Xiangpeng and Xu, Wenyuan and others},
  journal={arXiv preprint arXiv:2504.13914},
  year={2025}
}

@article{xu2025towards,
  title={Towards large reasoning models: A survey of reinforced reasoning with large language models},
  author={Xu, Fengli and Hao, Qianyue and Zong, Zefang and Wang, Jingwei and Zhang, Yunke and Wang, Jingyi and Lan, Xiaochong and Gong, Jiahui and Ouyang, Tianjian and Meng, Fanjin and others},
  journal={arXiv preprint arXiv:2501.09686},
  year={2025}
}

@inproceedings{wan2024alphazero,
  title={AlphaZero-Like Tree-Search can Guide Large Language Model Decoding and Training},
  author={Wan, Z and Feng, X and Wen, M and McAleer, SM and Wen, Y and Zhang, W and Wang, J},
  booktitle={Proceedings of Machine Learning Research},
  volume={235},
  pages={49890--49920},
  year={2024},
  organization={NeuriPS}
}

@inproceedings{hanposition,
  title={Position: General Intelligence Requires Reward-based Pretraining},
  author={Han, Seungwook and Pari, Jyothish and Gershman, Samuel J and Agrawal, Pulkit},
  booktitle={Forty-second International Conference on Machine Learning Position Paper Track},
  year={2025}
}

@article{han2025general,
  title={General reasoning requires learning to reason from the get-go},
  author={Han, Seungwook and Pari, Jyothish and Gershman, Samuel J and Agrawal, Pulkit},
  journal={arXiv preprint arXiv:2502.19402},
  year={2025}
}

@article{silver2016mastering,
  title={Mastering the game of Go with deep neural networks and tree search},
  author={Silver, David and Huang, Aja and Maddison, Chris J and Guez, Arthur and Sifre, Laurent and Van Den Driessche, George and Schrittwieser, Julian and Antonoglou, Ioannis and Panneershelvam, Veda and Lanctot, Marc and others},
  journal={nature},
  volume={529},
  number={7587},
  pages={484--489},
  year={2016},
  publisher={Nature Publishing Group}
}

@article{silver2017mastering,
  title={Mastering the game of go without human knowledge},
  author={Silver, David and Schrittwieser, Julian and Simonyan, Karen and Antonoglou, Ioannis and Huang, Aja and Guez, Arthur and Hubert, Thomas and Baker, Lucas and Lai, Matthew and Bolton, Adrian and others},
  journal={nature},
  volume={550},
  number={7676},
  pages={354--359},
  year={2017},
  publisher={Nature Publishing Group UK London}
}

@article{hao2024training,
  title={Training large language models to reason in a continuous latent space},
  author={Hao, Shibo and Sukhbaatar, Sainbayar and Su, DiJia and Li, Xian and Hu, Zhiting and Weston, Jason and Tian, Yuandong},
  journal={arXiv preprint arXiv:2412.06769},
  year={2024}
}

@article{shen2025codi,
  title={Codi: Compressing chain-of-thought into continuous space via self-distillation},
  author={Shen, Zhenyi and Yan, Hanqi and Zhang, Linhai and Hu, Zhanghao and Du, Yali and He, Yulan},
  journal={arXiv preprint arXiv:2502.21074},
  year={2025}
}

@article{zhang2025lightthinker,
  title={Lightthinker: Thinking step-by-step compression},
  author={Zhang, Jintian and Zhu, Yuqi and Sun, Mengshu and Luo, Yujie and Qiao, Shuofei and Du, Lun and Zheng, Da and Chen, Huajun and Zhang, Ningyu},
  journal={arXiv preprint arXiv:2502.15589},
  year={2025}
}

@article{aggarwal2025l1,
  title={L1: Controlling how long a reasoning model thinks with reinforcement learning},
  author={Aggarwal, Pranjal and Welleck, Sean},
  journal={arXiv preprint arXiv:2503.04697},
  year={2025}
}

@article{munkhbat2025self,
  title={Self-training elicits concise reasoning in large language models},
  author={Munkhbat, Tergel and Ho, Namgyu and Kim, Seo Hyun and Yang, Yongjin and Kim, Yujin and Yun, Se-Young},
  journal={arXiv preprint arXiv:2502.20122},
  year={2025}
}

@article{su2025between,
  title={Between underthinking and overthinking: An empirical study of reasoning length and correctness in llms},
  author={Su, Jinyan and Healey, Jennifer and Nakov, Preslav and Cardie, Claire},
  journal={arXiv preprint arXiv:2505.00127},
  year={2025}
}

@article{yi2025shorterbetter,
  title={Shorterbetter: Guiding reasoning models to find optimal inference length for efficient reasoning},
  author={Yi, Jingyang and Wang, Jiazheng and Li, Sida},
  journal={arXiv preprint arXiv:2504.21370},
  year={2025}
}

@article{yang2025towards,
  title={Towards thinking-optimal scaling of test-time compute for llm reasoning},
  author={Yang, Wenkai and Ma, Shuming and Lin, Yankai and Wei, Furu},
  journal={arXiv preprint arXiv:2502.18080},
  year={2025}
}

@article{sun2025empirical,
  title={An Empirical Study of LLM Reasoning Ability Under Strict Output Length Constraint},
  author={Sun, Yi and Wang, Han and Li, Jiaqiang and Liu, Jiacheng and Li, Xiangyu and Wen, Hao and Yuan, Yizhen and Zheng, Huiwen and Liang, Yan and Li, Yuanchun and others},
  journal={arXiv preprint arXiv:2504.14350},
  year={2025}
}

@article{zhang2025adaptthink,
  title={Adaptthink: Reasoning models can learn when to think},
  author={Zhang, Jiajie and Lin, Nianyi and Hou, Lei and Feng, Ling and Li, Juanzi},
  journal={arXiv preprint arXiv:2505.13417},
  year={2025}
}

@article{chen2024not,
  title={Do not think that much for 2+ 3=? on the overthinking of o1-like llms},
  author={Chen, Xingyu and Xu, Jiahao and Liang, Tian and He, Zhiwei and Pang, Jianhui and Yu, Dian and Song, Linfeng and Liu, Qiuzhi and Zhou, Mengfei and Zhang, Zhuosheng and others},
  journal={arXiv preprint arXiv:2412.21187},
  year={2024}
}

@inproceedings{qu2025optimizing,
  title={Optimizing Test-Time Compute via Meta Reinforcement Finetuning},
  author={Qu, Yuxiao and Yang, Matthew YR and Setlur, Amrith and Tunstall, Lewis and Beeching, Edward Emanuel and Salakhutdinov, Ruslan and Kumar, Aviral},
  booktitle={Forty-second International Conference on Machine Learning},
  year={2025}
}

@article{dai2025s,
  title={S-GRPO: Early Exit via Reinforcement Learning in Reasoning Models},
  author={Dai, Muzhi and Yang, Chenxu and Si, Qingyi},
  journal={arXiv preprint arXiv:2505.07686},
  year={2025}
}

@article{hiroshi2025practical,
  title={A Practical Two-Stage Recipe for Mathematical LLMs: Maximizing Accuracy with SFT and Efficiency with Reinforcement Learning},
  author={Hiroshi, Yoshihara and Taiki, Yamaguchi and Yuichi, Inoue},
  journal={arXiv preprint arXiv:2507.08267},
  year={2025}
}

@article{arora2025training,
  title={Training language models to reason efficiently},
  author={Arora, Daman and Zanette, Andrea},
  journal={arXiv preprint arXiv:2502.04463},
  year={2025}
}

@article{qi2025models,
  title={When Models Reason in Your Language: Controlling Thinking Trace Language Comes at the Cost of Accuracy},
  author={Qi, Jirui and Chen, Shan and Xiong, Zidi and Fern{\'a}ndez, Raquel and Bitterman, Danielle S and Bisazza, Arianna},
  journal={arXiv e-prints},
  pages={arXiv--2505},
  year={2025}
}

@article{wang2025thoughts,
  title={Thoughts are all over the place: On the underthinking of o1-like llms},
  author={Wang, Yue and Liu, Qiuzhi and Xu, Jiahao and Liang, Tian and Chen, Xingyu and He, Zhiwei and Song, Linfeng and Yu, Dian and Li, Juntao and Zhang, Zhuosheng and others},
  journal={arXiv preprint arXiv:2501.18585},
  year={2025}
}

@article{lanham2023measuring,
  title={Measuring faithfulness in chain-of-thought reasoning},
  author={Lanham, Tamera and Chen, Anna and Radhakrishnan, Ansh and Steiner, Benoit and Denison, Carson and Hernandez, Danny and Li, Dustin and Durmus, Esin and Hubinger, Evan and Kernion, Jackson and others},
  journal={arXiv preprint arXiv:2307.13702},
  year={2023}
}

@inproceedings{lyu2023faithful,
  title={Faithful Chain-of-Thought Reasoning},
  author={Lyu, Qing and Havaldar, Shreya and Stein, Adam and Zhang, Li and Rao, Delip and Wong, Eric and Apidianaki, Marianna and Callison-Burch, Chris},
  booktitle={Proceedings of the 13th International Joint Conference on Natural Language Processing and the 3rd Conference of the Asia-Pacific Chapter of the Association for Computational Linguistics (Volume 1: Long Papers)},
  pages={305--329},
  year={2023}
}

@inproceedings{ryu2024multi,
  title={Multi-Dimensional Optimization for Text Summarization via Reinforcement Learning},
  author={Ryu, Sangwon and Do, Heejin and Kim, Yunsu and Lee, Gary and Ok, Jungseul},
  booktitle={Proceedings of the 62nd Annual Meeting of the Association for Computational Linguistics (Volume 1: Long Papers)},
  pages={5858--5871},
  year={2024}
}

@article{mahan2024generative,
  title={Generative reward models},
  author={Mahan, Dakota and Van Phung, Duy and Rafailov, Rafael and Blagden, Chase and Lile, Nathan and Castricato, Louis and Fr{\"a}nken, Jan-Philipp and Finn, Chelsea and Albalak, Alon},
  journal={arXiv preprint arXiv:2410.12832},
  year={2024}
}

@article{yuan2024self,
  title={Self-rewarding language models},
  author={Yuan, Weizhe and Pang, Richard Yuanzhe and Cho, Kyunghyun and Sukhbaatar, Sainbayar and Xu, Jing and Weston, Jason},
  journal={arXiv preprint arXiv:2401.10020},
  volume={3},
  year={2024}
}

@article{wu2024meta,
  title={Meta-rewarding language models: Self-improving alignment with llm-as-a-meta-judge},
  author={Wu, Tianhao and Yuan, Weizhe and Golovneva, Olga and Xu, Jing and Tian, Yuandong and Jiao, Jiantao and Weston, Jason and Sukhbaatar, Sainbayar},
  journal={arXiv preprint arXiv:2407.19594},
  year={2024}
}

@inproceedings{huang2023large,
  title={Large Language Models Can Self-Improve},
  author={Huang, Jiaxin and Gu, Shixiang Shane and Hou, Le and Wu, Yuexin and Wang, Xuezhi and Yu, Hongkun and Han, Jiawei},
  booktitle={2023 Conference on Empirical Methods in Natural Language Processing, EMNLP 2023},
  pages={1051--1068},
  year={2023},
  organization={Association for Computational Linguistics (ACL)}
}

@article{zuo2025ttrl,
  title={Ttrl: Test-time reinforcement learning},
  author={Zuo, Yuxin and Zhang, Kaiyan and Sheng, Li and Qu, Shang and Cui, Ganqu and Zhu, Xuekai and Li, Haozhan and Zhang, Yuchen and Long, Xinwei and Hua, Ermo and others},
  journal={arXiv preprint arXiv:2504.16084},
  year={2025}
}

@article{zhang2025right,
  title={Right question is already half the answer: Fully unsupervised llm reasoning incentivization},
  author={Zhang, Qingyang and Wu, Haitao and Zhang, Changqing and Zhao, Peilin and Bian, Yatao},
  journal={arXiv preprint arXiv:2504.05812},
  year={2025}
}

@article{deepmath,
  title={DeepMath-103K: A Large-Scale, Challenging, Decontaminated, and  Verifiable Mathematical Dataset for Advancing Reasoning},
  author={He, Zhiwei and Liang, Tian and Xu, Jiahao and Liu, Qiuzhi and Chen, Xingyu and Wang, Yue and Song, Linfeng and Yu, Dian and Liang, Zhenwen and Wang, Wenxuan and Zhang, Zhuosheng and Wang, Rui and Tu, Zhaopeng and Mi, Haitao and Yu, Dong},
  year={2025},
  journal={arXiv preprint arXiv:2504.11456},
}

@inproceedings{lightman2023let,
  title={Let's verify step by step},
  author={Lightman, Hunter and Kosaraju, Vineet and Burda, Yuri and Edwards, Harrison and Baker, Bowen and Lee, Teddy and Leike, Jan and Schulman, John and Sutskever, Ilya and Cobbe, Karl},
  booktitle={The Twelfth International Conference on Learning Representations},
  year={2023}
}

@inproceedings{rein2024gpqa,
  title={Gpqa: A graduate-level google-proof q\&a benchmark},
  author={Rein, David and Hou, Betty Li and Stickland, Asa Cooper and Petty, Jackson and Pang, Richard Yuanzhe and Dirani, Julien and Michael, Julian and Bowman, Samuel R},
  booktitle={First Conference on Language Modeling},
  year={2024}
}

@article{clark2018think,
  title={Think you have solved question answering? try arc, the ai2 reasoning challenge},
  author={Clark, Peter and Cowhey, Isaac and Etzioni, Oren and Khot, Tushar and Sabharwal, Ashish and Schoenick, Carissa and Tafjord, Oyvind},
  journal={arXiv preprint arXiv:1803.05457},
  year={2018}
}

@article{wang2024mmlu,
  title={Mmlu-pro: A more robust and challenging multi-task language understanding benchmark},
  author={Wang, Yubo and Ma, Xueguang and Zhang, Ge and Ni, Yuansheng and Chandra, Abhranil and Guo, Shiguang and Ren, Weiming and Arulraj, Aaran and He, Xuan and Jiang, Ziyan and others},
  journal={Advances in Neural Information Processing Systems},
  volume={37},
  pages={95266--95290},
  year={2024}
}

@article{yu2025dapo,
  title={Dapo: An open-source llm reinforcement learning system at scale},
  author={Yu, Qiying and Zhang, Zheng and Zhu, Ruofei and Yuan, Yufeng and Zuo, Xiaochen and Yue, Yu and Dai, Weinan and Fan, Tiantian and Liu, Gaohong and Liu, Lingjun and others},
  journal={arXiv preprint arXiv:2503.14476},
  year={2025}
}

@article{meng2024simpo,
  title={Simpo: Simple preference optimization with a reference-free reward},
  author={Meng, Yu and Xia, Mengzhou and Chen, Danqi},
  journal={Advances in Neural Information Processing Systems},
  volume={37},
  pages={124198--124235},
  year={2024}
}

\clearpage
\appendix
\setcounter{secnumdepth}{2}

\section{Advantage Computation}
\label{app:advantage}

As delineated in Section \ref{sec:method} and illustrated in Figure \ref{fig:method}, the reward computation for a given response relies on two factors, namely (1) its correctness and (2) its status as a rewritten version. An incorrect response is always assigned a zero reward. Conversely, a correct rewritten response is assigned the full reward of $1.0$, which aligns with intuitive expectations. However, a correct, non-rewritten response is also assigned a zero reward, identical to an incorrect response. This equivalency may raise concerns, as conventional approaches would typically assign a non-zero, fractional reward (e.g., $0 < \text{reward} < 1$) to a correct non-rewritten output.

The rationale for this reward structure lies in the GRPO normalization applied during the advantage computation. Given the selective rewriting mechanism, only the correct non-rewritten responses and the correct rewritten responses are grouped together for advantage calculation and subsequent normalization. Within this specific grouping, assigning any positive, constant reward to correct, non-rewritten responses yields an identical outcome after normalization via the group's standard deviation. Therefore, assigning a zero reward or any other constant positive reward to a correct, non-rewritten response has the same effective impact on the computed advantage. For pragmatic simplicity, we elect to use a zero reward.

To conceptually incorporate the intuition of preferring correct non-rewritten responses over incorrect ones without compromising the advantage normalization, we introduce a scaling factor to the advantages. Specifically, all advantages within a rewritten batch are divided by $5$. By reducing the magnitude of these advantages, we constrain the policy's preference for correct rewritten responses over correct non-rewritten ones, thereby implicitly granting the latter a form of credit. Operationally, dividing the advantages by $5$ is mathematically equivalent to setting the learning rate for all samples involved in the rewriting process to $1/5$.

\section{Prompts}
\label{app:prompts}

\subsection{Generation}

\begin{tcolorbox}[
    colback=gray!10,
    colframe=gray!75!black,
    title={Math Questions},
]
Solve the following math problem. Put your final answer within \textbackslash boxed\{\}.\\ \\
\{query\}
\end{tcolorbox}

\begin{tcolorbox}[
    breakable,
    colback=gray!10,
    colframe=gray!75!black,
    title={TOPS Low Effort Math Questions},
]
You have extremely limited time to think and respond to the user's query. Every additional second of processing and reasoning incurs a significant resource cost, which could affect efficiency and effectiveness. Your task is to prioritize speed without sacrificing essential clarity or accuracy. Provide the most direct and concise answer possible. Avoid unnecessary steps, reflections, verification, or refinements UNLESS ABSOLUTELY NECESSARY. Your primary goal is to deliver a quick, clear and correct response. Put your final answer within \textbackslash boxed\{\}.\\ \\
\{query\}
\end{tcolorbox}

\begin{tcolorbox}[
    colback=gray!10,
    colframe=gray!75!black,
    title={TOPS High Effort Math Questions},
]
You have unlimited time to think and respond to the user’s question. There is no need to worry about reasoning time or associated costs. Your only goal is to arrive at a reliable, correct final answer. Feel free to explore the problem from multiple angles, and try various methods in your reasoning. This includes reflecting on reasoning by trying different approaches, verifying steps from different aspects, and rethinking your conclusions as needed. You are encouraged to take the time to analyze the problem thoroughly, reflect on your reasoning promptly and test all possible solutions. Only
after a deep, comprehensive thought process should you provide the final answer, ensuring it is correct and well-supported by your reasoning. Put your final answer within \textbackslash boxed\{\}.\\ \\
\{query\}
\end{tcolorbox}

\begin{tcolorbox}[
    float,
    breakable,
    colback=gray!10,
    colframe=gray!75!black,
    title={Multi-Choice Questions},
]
Select one option from the given choices that accurately answers the question, and enclose your final answer (which should be the uppercase letter index of the chosen option) within \textbackslash boxed\{\}.\\ \\
\#\#\# Question \\
\{question\} \\ \\
\#\#\# Choices \\
\{choices\}
\end{tcolorbox}

\subsection{Rewriting}

\begin{tcolorbox}[
    breakable,
    colback=gray!10,
    colframe=gray!75!black,
]
\#\#\# Instruction\\
You are a skilled editor tasked with improving a given thinking passage. Your goal is to refine the passage to enhance its overall quality, making it more organized, coherent, and accurate. Your output should be a rewritten version of the original thinking passage. The rewritten version should maintain the core ideas and essence of the original while significantly improving its presentation and impact. Note that:\\ \\
1. Always use a first-person tone when refining.\\
2. This is more like thinking out loud than proper writing. Use simple and everyday language.\\
3. It's okay, and even good, to use sentences of reflecting, pausing to think, or connecting different thoughts. But if such verbose sentences bring no significant new ideas, you may simplify or remove them for clarity.\\
4. Stick to the format and language style of the original.\\
Please provide only the rewritten thinking passage, without any additional explanations or context.\\ \\
\#\#\# Thinking Passage to Rewrite \\
\{passage\}
\end{tcolorbox}

\subsection{Judging}

\begin{tcolorbox}[
    breakable,
    colback=gray!10,
    colframe=gray!75!black,
]
\#\#\# Instruction\\
You are an impartial judge evaluating the overall quality of a piece of thinking texts. Your task is to assess the provided texts considering the following flaws.\\ \\
1.  **Over-Thinking**: The passage exhibits excessive thinking on simple or straightforward concepts.\\
2.  **Under-Thinking**: The passage lacks sufficient depth, complexity, or thoroughness in addressing significant and challenging aspects.\\
3.  **Disordered-Thinking**: The flow of thought is illogical and inconsistent, or interleaves multiple unrelated topics, making it difficult to follow.\\
4.  **Redundant-Thinking**: The passage repeats ideas or insights unnecessarily without adding significant new value or perspective.\\ \\
For each flaw, assign a score from 1 to 5, where 1 indicates the flaw is significantly present (worst) and 5 indicates the flaw is perfectly avoided (best). After evaluating each aspect, provide an overall judgment of the passage's quality, also on a 1-5 scale. Your output must strictly adhere to the following format, where `[Your judgment]' is a single numerical digit (1, 2, 3, 4, or 5):\\ \\
\textasciigrave\textasciigrave\textasciigrave\\
Aspect 1: [Your judgment]\\
Aspect 2: [Your judgment]\\
Aspect 3: [Your judgment]\\
Aspect 4: [Your judgment]\\
Overall: [Your judgment]\\
\textasciigrave\textasciigrave\textasciigrave\\ \\
Please provide only the formatted judgment, without any additional explanations or context outside of the specified format.\\ \\
\#\#\# Thinking Passage to Judge\\
\{passage\}
\end{tcolorbox}

\section{Experimental Details}
\label{app:implementation}

\subsection{Training Framework and Settings}

\paragraph{Online RL}
The online GRPO-based methods including vanilla GRPO, LenPen1, LenPen2, ShorterBetter, and ours self-rewriting are implemented based on \texttt{verl}\footnote{\url{https://github.com/volcengine/verl}} framework. We adopt parameter offloading, optimizer offloading, and gradient checkpoint techniques for memory efficiency. The online inference utlize \texttt{vllm}\footnote{\url{https://github.com/vllm-project/vllm}} framework.

\paragraph{Offline RL}
The offline SimPO methods including LPO and TOPS are implemented based on \texttt{LLaMA-Factory}\footnote{\url{https://github.com/hiyouga/LLaMA-Factory}} framework. We adopt parameter offloading, optimizer offloading, and gradient checkpoint techniques for memory efficiency.

\subsection{Hyper-Parameters}

The hyper-parameters are listed in Table \ref{tab:training-args} and \ref{tab:inference-args} respectively.

\begin{table}[htb!]
    \centering
    \begin{tabular}{lcc}
        \toprule
        \textbf{Hyper-parameter} & \textbf{Online} & \textbf{Offline}\\
        \midrule
        parameters & full & full \\
        epoch & $1$ & $1$ \\
        batch size & $256$ & $256$ \\
        learning rate & $3e-6$ & $1e-6$ \\
        learning rate shceduler & constant & constant \\
        gradient norm & $1.0$ & $1.0$ \\
        optimizer & AdamW & AdamW \\
        weight decay & $1e-2$ & $1e-2$ \\
        cutoff length & $12288$ & $12288$ \\

        rollout size & $8$ & - \\
        PPO $\epsilon$ & $0.2$ & - \\
        KL loss coefficient & $0.0$ & - \\

        preference $\beta$ & - & $2.0$ \\
        simpo $\gamma$ & - & $0.3$ \\
        \bottomrule
    \end{tabular}
    \caption{Hyper-parameters for training.}
    \label{tab:training-args}
\end{table}

\begin{table}[htb!]
    \centering
    \begin{tabular}{lcc}
        \toprule
        \textbf{Hyper-parameter} & \textbf{Sample} & \textbf{Rewrite and Test}\\
        \midrule
        max new tokens & $12288$ & $32768$ \\
        temperature & $1.0$ & $0.6$ \\
        top k & $-1$ & $20$ \\
        top p & $1.0$ & $0.95$ \\
        \bottomrule
    \end{tabular}
    \caption{Hyper-parameters for inference.}
    \label{tab:inference-args}
\end{table}

\begin{table*}[htb!]
    \centering
    \setlength{\tabcolsep}{1mm}
    \begin{tabular}{lcccccccccccc|ccc}
        \toprule
        \multirow{2}{*}{\textbf{Models}} & \multicolumn{3}{c}{\textbf{MATH-500}} & \multicolumn{3}{c}{\textbf{GPQA-Diamond}} & \multicolumn{3}{c}{\textbf{ARC-Challenge}} & \multicolumn{3}{c}{\textbf{MMLU-Pro}} & \multicolumn{3}{|c}{\textbf{Average}} \\
        \cmidrule(lr){2 - 4} \cmidrule(lr){5 - 7} \cmidrule(lr){8 - 10} \cmidrule(lr){11 - 13} \cmidrule(lr){14 - 16} & \textbf{GPT} & \textbf{DB} & \textbf{DS} & \textbf{GPT} & \textbf{DB} & \textbf{DS} & \textbf{GPT} & \textbf{DB} & \textbf{DS} & \textbf{GPT} & \textbf{DB} & \textbf{DS} & \textbf{GPT} & \textbf{DB} & \textbf{DS} \\
        \midrule
        Original &
        45.1 & 54.1 & 78.6 &
        31.9 & 39.1 & 58.5 &
        58.7 & 70.8 & 82.1 &
        41.0 & 53.3 & 69.3 &
        44.2(+0.0) & 54.3(+0.0) & 72.1(+0.0) \\
        LenPen1 &
        50.2 & 63.0 & 83.6 &
        34.0 & 42.5 & 58.8 &
        63.1 & 74.6 & 85.4 &
        46.9 & 59.3 & 72.1 &
        48.5(+4.3) & 59.3(+5.0) & 75.0(+2.9) \\
        LenPen2 &
        51.0 & 63.5 & 82.9 &
        33.8 & 41.7 & 59.0 &
        64.5 & 74.8 & 84.5 &
        46.6 & 59.6 & 73.5 &
        48.9(+4.7) & 59.9(+5.6) & 75.0(+2.9) \\
        Rewrite           & 
        55.4 & 69.7 & 90.0 &
        36.3 & 45.9 & 61.2 &
        68.6 & 80.4 & 89.5 &
        48.7 & 62.5 & 76.6 &
        52.2(+8.0) & 64.6(+10.3) & 79.3(+7.2) \\
        \bottomrule
    \end{tabular}
    \caption{LLM judge scores on Qwen3-8B-based models. GPT, DB, DS stand for GPT-4o, Doubao-1.5-Lite, and DeepSeek-V3 respectively.}
    \label{tab:additional-judges}
\end{table*}

\subsection{Evaluation}

\paragraph{Math Questions}
Following previous work, MATH-500 is evaluated with the script of original MATH dataset\footnote{\url{https://github.com/hendrycks/math/blob/main/modeling/math_equivalence.py}}. While during training, DeepMath-103K is evaluated with \texttt{Math-Verify}\footnote{\url{https://github.com/huggingface/Math-Verify}} library.

\paragraph{Multi-Choice Questions}
Tasks of GPQA-Diamond, ARC-Challenge, and MMLU-Pro present multi-choice questions where the task is to choose one option from the given ones. An answer is considered correct if it exactly matches the golden label.

\section{Other LLM Judges}
\label{app:other-judges}

Additional results using other LLM judges are shown in Table \ref{tab:additional-judges}. Under all judges, our proposed self-rewriting exhibits stronger reasoning capability.

\section{Case Study}

A case of original and rewritten reasoning is shown in the boxes below. The internal reasoning flaws are marked, including:
\begin{itemize}
    \item \texttt{Over1} Over-thiniking about the chemical equation, which is actually unnecessary for problem solving.
    \item \texttt{Over2} Over-thinking about the molar masses.
    \item \texttt{Redundant1} Repeating the discussion about the container air tightness.
    \item \texttt{Redundant2} Repeating the chemical equation without new information.
    \item \texttt{Disordered1} Interleaved thoughts about the chemical equation and the total mass.
    \item \texttt{Disordered2} Interleaved thoughts about the total mass and air tightness.
    \item \texttt{Disordered3} Interleaved thoughts about air tightness and the final answer.
\end{itemize}

The comparison showcases that rewriting can significantly improving internal reasoning quality by reducing redundancy and maintaining consistency, validating the LLM-as-a-judge score improvement.

\begin{tcolorbox}[
    float*,
    width=\textwidth,
    breakable,
    colback=gray!10,
    colframe=gray!75!black,
    title={Original Reasoning},
]
Okay, so I need to figure out the final mass of the products when 100 grams of vinegar and 5 grams of baking soda react, and the gas is trapped in the container. Let me think about this step by step.\\ \\
\texttt{[Over1]} First, I remember that when vinegar (which is acetic acid) reacts with baking soda (sodium bicarbonate), they produce carbon dioxide gas, water, and sodium acetate. The chemical equation for this reaction is:\\ \\
\texttt{[Over1]} CH3COOH (vinegar) + NaHCO3 (baking soda) $\rightarrow$ CO2 (gas) + H2O + CH3COONa (sodium acetate)\\ \\
\texttt{[Over1, Disordered1]} So, the reactants are vinegar and baking soda, and the products are carbon dioxide, water, and sodium acetate. But the question says the gas is trapped in the container. That means the gas isn't escaping, so it's part of the total mass.\\ \\
Now, the question is about the final mass of the products. According to the law of conservation of mass, mass should be conserved in a closed system. So the total mass before and after the reaction should be the same, right? But wait, the problem says the gas is trapped, so it's not lost. Therefore, the total mass of the products should equal the total mass of the reactants.\\ \\
\texttt{[Disordered2]} Wait, the initial mass is 100 grams of vinegar plus 5 grams of baking soda, which totals 105 grams. If the gas is trapped, then all the products are accounted for, so the final mass should be 105 grams. But let me check the options. Choice C is 105 grams. But wait, maybe I need to consider the gas?\\ \\
\texttt{[Redundant1]} Wait, no. Because the gas is part of the products. So the total mass of the products is the mass of the solid products (sodium acetate and water) plus the trapped gas. But the original mass of the reactants is 100 + 5 = 105 grams. So the total mass should remain 105 grams. Therefore, the answer would be (C) 105 grams.\\ \\
\texttt{[Redundant1, Disordered3]} But wait, sometimes in reactions, if a gas is produced, people might think that the mass decreases if the gas is allowed to escape. But in this case, the gas is trapped, so it's included in the total mass. Therefore, the total mass remains the same as the initial mass. So the final mass should be 105 grams. That's option C. Let me check the options again. The choices are (A) 100, (B) 104, (C) 105, (D) 110. So C is 105 grams. Therefore, the answer should be C.\\ \\
\texttt{[Over1, Redundant2]} Wait, but maybe there's a trick here. Let me think again. Let me write down the reaction properly. The balanced equation is:\\ \\
\texttt{[Over1, Redundant2]} CH3COOH + NaHCO3 $\rightarrow$ CO2 + H2O + CH3COONa\\ \\
\texttt{[Over2]} So, the molar masses: Let me check the stoichiometry. Let's say 1 mole of acetic acid reacts with 1 mole of sodium bicarbonate to produce 1 mole of CO2, 1 mole of H2O, and 1 mole of sodium acetate.\\ \\
\texttt{[Redundant1]} So, the total mass of the products would be the sum of the masses of CO2, H2O, and sodium acetate. But the initial mass is vinegar (acetic acid) and baking soda. But since the reaction is between them, the total mass of the products should be equal to the sum of the masses of the reactants, because the gas is trapped. Therefore, the total mass remains 100 + 5 = 105 grams. Therefore, the answer is C.\\ \\
\texttt{[Redundant1]} But wait, another thought: if the gas is trapped, then the mass of the container plus the gas would be the same as the initial mass. So, the answer is C. So, I think the answer is C, 105 grams. Let me check if there's any other factor. For example, sometimes when reactions occur, if there's a gas, but in this case, the gas is trapped. Therefore, the mass should be conserved. So, the answer is C. So the correct choice is C.\\ \\
**Final Answer**\\
\boxed{C}
\end{tcolorbox}

\begin{tcolorbox}[
    float*,
    width=\textwidth,
    breakable,
    colback=gray!10,
    colframe=gray!75!black,
    title={Rewritten Reasoning},
]
Okay, so I need to figure out the final mass of the products when 100 grams of vinegar and 5 grams of baking soda react, and the gas is trapped in the container. Let me start by recalling what happens in this reaction.\\ \\
\texttt{[Over1]} CH3COOH + NaHCO3 $\rightarrow$ CO2 + H2O + CH3COONa\\ \\
\texttt{[Over1, Disordered1]} So the reactants are vinegar and baking soda, and the products are carbon dioxide, water, and sodium acetate. But the question says the gas is trapped in the container. That means the gas isn't escaping, so it's part of the total mass.\\ \\
Now, the question is about the final mass of the products. I remember the law of conservation of mass says mass is conserved in a closed system. So the total mass before and after the reaction should be the same. But wait, the problem says the gas is trapped, so it's not lost. Therefore, the total mass of the products should equal the total mass of the reactants.\\ \\
The initial mass is 100 grams of vinegar plus 5 grams of baking soda, which totals 105 grams. If the gas is trapped, then all the products are accounted for, so the final mass should be 105 grams. That would be option C.\\ \\
\texttt{[Redundant 1]} But wait, maybe I need to consider the gas again. The gas is part of the products, so the total mass of the products includes the gas. Since the gas is trapped, it doesn't escape, so the total mass remains the same as the initial mass. Therefore, the final mass should be 105 grams.\\ \\
Let me check the options again. The choices are (A) 100, (B) 104, (C) 105, (D) 110. So C is 105 grams. That makes sense.\\ \\
\texttt{[Redundant 1]} Another thought: sometimes people might think the mass decreases if gas is produced, but in this case, the gas is trapped, so it's included. Therefore, the total mass stays the same. So the answer is C.\\ \\
I think that's it. The answer should be C. Let me just make sure I didn't miss anything. The reaction is balanced, the gas is trapped, so mass is conserved. Yep, that's solid. The final mass is 105 grams.\\ \\
**Final Answer**\\
\boxed{C}
\end{tcolorbox}

\end{document}